\documentclass[letterpaper,10pt]{article} 

\usepackage{opticameet3} 

\usepackage{array}

\usepackage{makecell}

\usepackage{booktabs}

\newcommand\authormark[1]{\textsuperscript{#1}}

\usepackage{amsmath,amssymb}
\usepackage[colorlinks=true,bookmarks=false,citecolor=blue,urlcolor=blue]{hyperref} 

\usepackage[backend=biber,style=authoryear, sorting=nyt]{biblatex} 
\usepackage{bbding}

\usepackage{tabularx}

\usepackage{graphicx}
\graphicspath{ {images/} }

\DeclareUnicodeCharacter{0301}{\'{e}}

\usepackage{titlesec}

\titleformat{\paragraph}
{\normalfont\small}{\theparagraph}{1em}{}
\titlespacing*{\paragraph}
{0pt}{3.25ex plus 1ex minus .2ex}{1.5ex plus .2ex}

\usepackage{lineno}

\usepackage{setspace}

\providecommand{\keywords}
{
\textbf{\text{Keywords:}}
}

\begin{document}
This paper has been submitted to Computers and Electronics in Agriculture for review.
\newpage

\title{A Comprehensive Review on Tree Detection Methods Using Point Cloud and Aerial Imagery from Unmanned Aerial Vehicles}


\author{Weijie Kuang\authormark{1}, Hann Woei Ho\authormark{1,*}, Ye Zhou\authormark{1}, Shahrel Azmin Suandi\authormark{2}, and \\Farzad Ismail\authormark{1}}

\address{\authormark{1} School of Aerospace Engineering, Engineering Campus, Universiti Sains Malaysia, 14300 Nibong Tebal, Pulau Pinang, Malaysia\\
\authormark{2}School of Electrical and Electronic Engineering, Engineering Campus, Universiti Sains Malaysia, 14300 Nibong Tebal, Pulau Pinang, Malaysia}

\email{\authormark{*}Corresponding author. Email: aehannwoei@usm.my} 

\begin{doublespace}

\begin{abstract}

Unmanned Aerial Vehicles (UAVs) are considered cutting-edge technology with highly cost-effective and flexible usage scenarios. Although many papers have reviewed the application of UAVs in agriculture, the review of the application for tree detection is still insufficient. This paper focuses on tree detection methods applied to UAV data collected by UAVs. There are two kinds of data, the point cloud and the images, which are acquired by the Light Detection and Ranging (LiDAR) sensor and camera, respectively. Among the detection methods using point-cloud data, this paper mainly classifies these methods according to LiDAR and Digital Aerial Photography (DAP). For the detection methods using images directly, this paper reviews these methods by whether or not to use the Deep Learning (DL) method. Our review concludes and analyses the comparison and combination between the application of LiDAR-based and DAP-based point cloud data. The performance, relative merits, and application fields of the methods are also introduced.
Meanwhile, this review counts the number of tree detection studies using different methods in recent years. From our statics, the detection task using DL methods on the image has become a mainstream trend as the number of DL-based detection researches increases to 45 \% of the total number of tree detection studies up to 2022. As a result, this review could help and guide researchers who want to carry out tree detection on specific forests and for farmers to use UAVs in managing agriculture production.

\end{abstract}

\section{Introduction}
Forest plays a crucial role in the global ecosystem. Protecting and managing forests is essential for the environment and sustainable development. In the management of forests, the Individual Tree Detection (ITD) could provide data support in biomass estimation, growing situation monitoring, and precise spraying \parencite{mohan2017individual}. Traditional tree detection relies on the field data and statistical estimators. \parencite{cabrera2022individualization}. However, the conventional forest inventory, as field measurements, is time-intensive and expensive to get accurate data from small-region plantations \parencite{hyyppa2001segmentation}. Therefore, highly accurate, cost-effective, rapid forest inventory techniques are needed to enable tree detection to address the increasing demand for efficient forestry management. Although remote sensing images from satellites have been used for acquiring forest information for many years, the low quality of the freely available satellite images or expensive high resolution images restricts its application in the small-scale forest management \parencite{white2016remote}.


 Unmanned Aerial Vehicles (UAVs), originally developed for military purposes as cutting-edge technology, are now steadily making their way into commercial applications within the agroforestry sector \parencite{hu2020development}. UAVs offer distinct advantages over traditional remote sensing methods, primarily owing to their various designs \parencite{wong2021design}, making them versatile tools in this context. Compared to other forms of remote sensing, UAVs are capable of acquiring image data with minimal atmospheric interference, at a lower cost, with greater flexibility in usage timing, and boasting higher spatial resolution. Consequently, UAV-captured remote sensing data is particularly well-suited for agroforestry applications, especially in smaller plantations. In the realm of tree detection, two primary models emerge: point cloud-based and spectral image-based detection. Notably, point cloud-based detection leverages data collected by Light Detection And Ranging (LiDAR) sensors or cameras mounted on UAVs, showcasing the adaptability and potential for customization inherent in UAV designs.

LiDAR could determine the distance of objects by measuring the interval time between the laser pulse and the reflected signal without being affected by the weather \parencite{hui2022multi}. The developments of the flight control, the motor and the battery on UAVs enhance the ability of LiDAR to collect three-dimensional structure data of forests. Besides, LiDAR could detect the terrain information of the forest floor directly due to the penetration of laser beam on forest canopy \parencite{rahman2022forest}. However, the high cost of LiDAR makes it challenging to be used in small area forests universally. 
Meanwhile, the emergence of image stereomatching makes the reconstruction of the 3D structure of trees possible in the field of Digital Aerial Photography (DAP) \parencite{mielcarek2020digital}. Among them, Structure from Motion (SfM) is a common technique reproduces the 3D structure by processing the two-dimensional overlapping photos which target on the same object from different directions. Generally, the tree detection approaches use the point cloud data in two steps, i.e., tree localization and segmentation. For example, to find the location of the tree top, a Local Maximum (LM) algorithm is first applied on Canopy Height Models (CHM) which represents the residual distance between the ground and the object top \parencite{popescu2002estimating}. Then, a segmentation algorithm, Watershed Algorithm (WA), is applied to separate an individual tree from other trees, or background \parencite{yin2020individual}. Finally, the ITD is achieved with the localization of the tree top and the segmentation of the tree crown area.

Moreover, the LM was applied on the CHM In detecting the treetop in mountainous forests. The results showed that 85 \% of the treetop could be detected when the UAV was flown following terrain \parencite{gonroudobou2022treetop}. In the ITD of the crown over pure ginkgo, one approach used the WA on the CHM obtained from LiDAR for individual tree segmentation, and then the LM was used to localize the tree top. However, the detection using WA did not show good results since the WA often divides a single tree into several trees \parencite{wu2019assessment}. To evaluate and compare the performance of LiDAR-based and DAP-based point cloud data on Eucaluptus tree detection, another study also implemented LM with a fixed window size \parencite{guerra2018comparison}. The results showed that the UAV images had a similar capability with LiDAR-based point clouds on the ITD. Therefore, point clouds generated from UAV images could be a robust and low-cost alternative for expensive laser scanning sensors to be used in tree detection. Usually, the tree detection is completed by combining several methods, such as Point Cloud-Based Cluster Segmentation, Layer Stacking, and k-means clustering, which will be introduced in the following parts \parencite{ma2022performance}. However, the points derived from aerial images do not have good qualities in the alpine forests. Because the irregular topography makes the aerial images taken inconsistent. And the high tree density hinders the reconstruction of the single tree. 

Rather than detecting trees on the point cloud data, with the advances of computer vision technology, more and more researchers devoted to conducting the detection task on the UAV image solely \parencite{adao2017hyperspectral, diez2021deep}. The methods used in tree detection could be divided into classical and Deep Learning (DL) methods. 
The classical methods, such as support vector machine, scale-invariant feature transform, superpixel, optical flow, and morphological operation, often do the detection task based on the pixel level. For instance, in the automated delineation of crop crown research, researchers used morphological operations to segment the UAV images to separate the trees from each other \parencite{ponce2021methodology}. Besides static images, by detecting and tracking features in consecutive images, optical flow representing the relative motion to detected features can be obtained for identifying objects underneath UAVs \cite{ho2015optical, ho2016characterization}.
On the other hand, DL methods usually use multiple networks to detect trees based on the object \parencite{ching2022ultra, lee2021deep}. Furthermore, many DL methods take candidate boxes into use and then give them scores toward different classes \parencite{osco2021review}. For instance, object detection DL algorithms, such as Faster-RCNN and YOLO-V4, were applied to realise the recognition of the tree with the pine wilt disease. Results showed that the Faster R-CNN has a higher detection accuracy of 66.7 \% than the YOLO-V4 as 63.55 \% \parencite{yu2021early}. Apart from the object-based DL methods, DL networks are also focused on scoring each pixel. For example, one study used the convolutional neural networks (CNN) which is based on the shared-weight architecture of the convolution kernels to identify citrus trees on UAV images \parencite{csillik2018identification}. At first, the CNN will generate the heat map to show the probability that each pixel belongs to the tree. To reduce the error induced by the multiple detection of large canopies during the CNN, simple linear iterative clustering is then applied to the heat map to aggregate all the objects representing the same tree.

\begin{figure}[ht]
\centering
  \includegraphics[height=6.2cm]{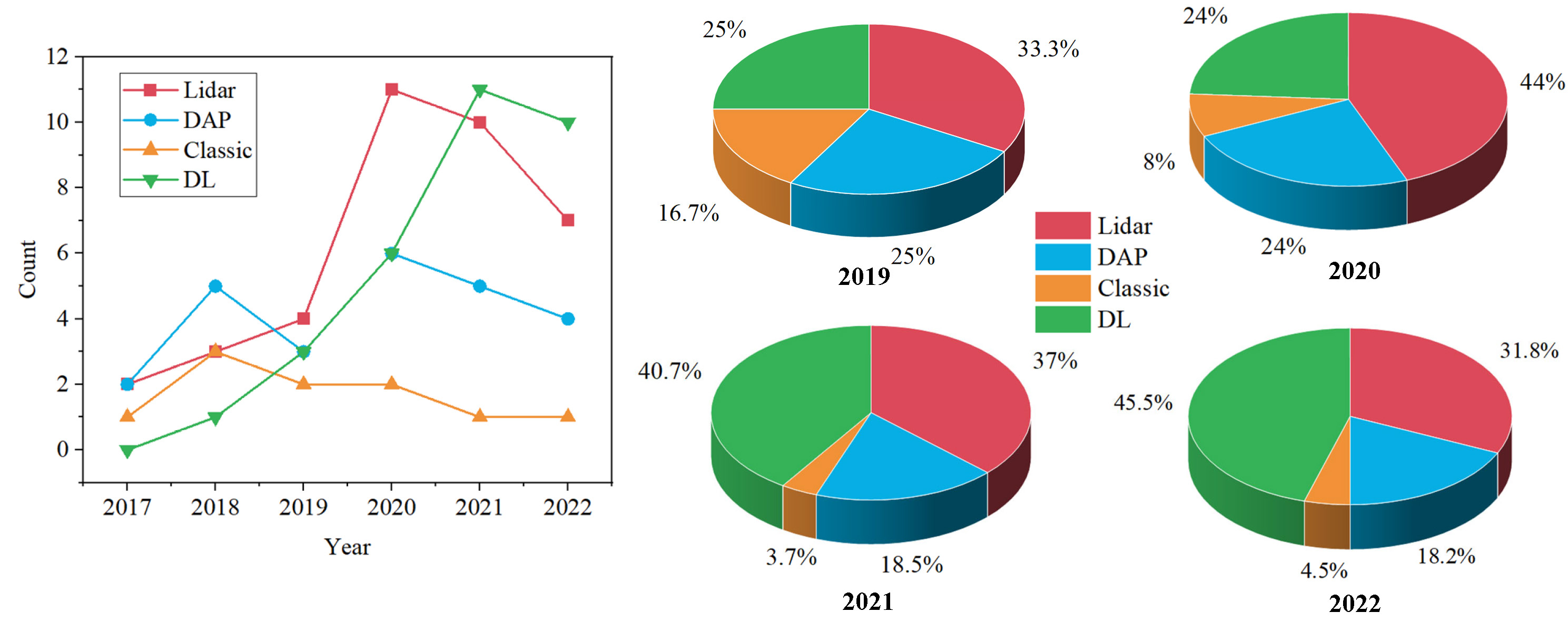}
  \caption{Quantity of articles published on tree detection in recent years }
  \label{fig:Quantity}
\end{figure}

Fig. \ref{fig:Quantity} shows the number of articles published on tree detection using UAVs in recent years. All articles were counted according to the following four categories, which have been explained in the above content detailedly: (1) LiDAR represents the research that uses LiDAR sensors to collect the point cloud data; (2) DAP represents the research that generates the point cloud data from UAV images; (3) Classic represent the researches that apply the classical methods on UAV images directly; (4) DL represent the researches that use the DL methods on UAV images directly. As can be seen from the line chart in \ref{fig:Quantity}, the tree detection on LiDAR and DL make up the majority. Moreover, the quantity of LiDAR-based tree detection increased continuously from 2017 to 2020 whereas it began to reduce from 2020 to 2022.

Furthermore, the total number of studies using point cloud data for tree detection is decreasing, but the number of studies using LiDAR data has remained significant. This is due to the high precision of the point cloud data collected by LiDAR sensor, even if it is more expensive than other methods. For DL-based tree detection, the number of it has increased dramatically due to the significant progress made in object detection algorithms in the DL research area \parencite{kaur2022comprehensive}. Meanwhile, the number of research on classic-based tree detection remains the lowest. It decreases every year caused by the high dependency of handcraft methods on hyperparameters, which are usually set manually and restricted to a specific application. The pie chart shows that the research on DL-based tree detection increases to 45.5 \% of the total number of studies up to 2022. Moreover, even though the percentage of LiDAR-based tree detection decreases, it is still one of the main directions of tree detection research.

Recently, many researchers have reviewed the application of UAVs in forestry, agriculture, and agroforestry, as summarised in Table. \ref{table:summary}. This summary includes the year, keywords, does the article contain tree detection, and published papers. Even though some reviews include tree detection, it is only a small supporting part, not its subject matter. Therefore, to the best of our knowledge, there is a literature gap related to review articles combining the application of UAVs and tree detection. The main goal of this review is to analyze, conclude, and look to the future of the application of UAVs, as a rapidly growing research field, on tree detection. Our review aims to introduce existing methods and common shortcomings. In short, the contributions of this paper are the following:

\begin{itemize}
\item The statistics on published articles in scientific sources according to the publication year, data type and method type are provided. And the statistical results on the development trend of UAV application on tree detection are analysed. 
\item The presentation of the principle, application scenario and shortcomings of existing tree detection methods is given. And the solutions for the shortcoming of the earlier methods are introduced.
\item The challenge of tree detection conducted by the UAV systems in the complicated environment are described. The possible solutions and future perspectives are offered.
\end{itemize}

\begin{table}[htb]
\label{table:summary}
 \centering \caption{Summary of reviews relating to forestry, agriculture, and agroforestry}
 
 \begin{tabularx}{\textwidth} { m{1.1cm}<{\centering} m{1cm}<{\centering} m{6cm}<{\centering} m{1.5cm}<{\centering} m{5cm}<{\centering}}
    \Xhline{1.5pt}
    Reference & Year & Keywords & Tree detection & Journal \\
    \Xhline{1pt}
    
    \parencite{adao2017hyperspectral} & 2017 & Hyperspectral; UAS; UAV; Hyperspectral sensors; Hyperspectral data processing; Agriculture; Forestry; Agroforestry & \XSolidBrush & Remote sensing  \\
    
     \parencite{xie2020review} & 2020 & Unmanned aerial system (UAS); Sensors; Plant; Phenotyping; Traits & \XSolidBrush & Computers and Electronics in Agriculture  \\

    \parencite{hu2020development} & 2020 & Unmanned Aerial Vehicle (UAV); Lidar; DJI Livox; Low cost; Forest inventory & \Checkmark & Remote Sensing  \\
   
    \parencite{feng2021comprehensive} & 2021 & Unmanned aerial vehicle; Remote sensing; High-throughput phenotyping; Sensors; Applications review & \XSolidBrush & Computers and Electronics in Agriculture  \\
   
    \parencite{osco2021review} & 2021 & Convolutional neural networks; Remote sensing imagery; Unmanned aerial vehicles & \Checkmark & International Journal of Applied Earth Observations and Geoinformation  \\

    \parencite{diez2021deep} & 2021 & Deep learning; UAV; Forestry; Literature review; Practical applications; RGB & \Checkmark & Remote Sensing  \\
   
    \parencite{albiero2022swarm} & 2022 & Operational cost; Electric tractor; Artificial intelligence; Plowing; Forestry; Multi-robot & \XSolidBrush & Computers and Electronics in Agriculture  \\

     \parencite{pathak2022review} & 2022 & Computer vision; Deep learning; Image processing; Machine learning; Machine vision; Plant stand count & \Checkmark & Computers and Electronics in Agriculture  \\

      \parencite{rejeb2022drones} & 2022 & Drones; UAV; Precision agriculture; Internet of Things; Bibliometric & \XSolidBrush & Computers and Electronics in Agriculture  \\
      
    \Xhline{1.5pt}
   \end{tabularx}
  
    \end{table}

\section{Object Detection based on 3D Point Cloud}
\begin{figure}[ht]
\centering
  \includegraphics[height=9cm]{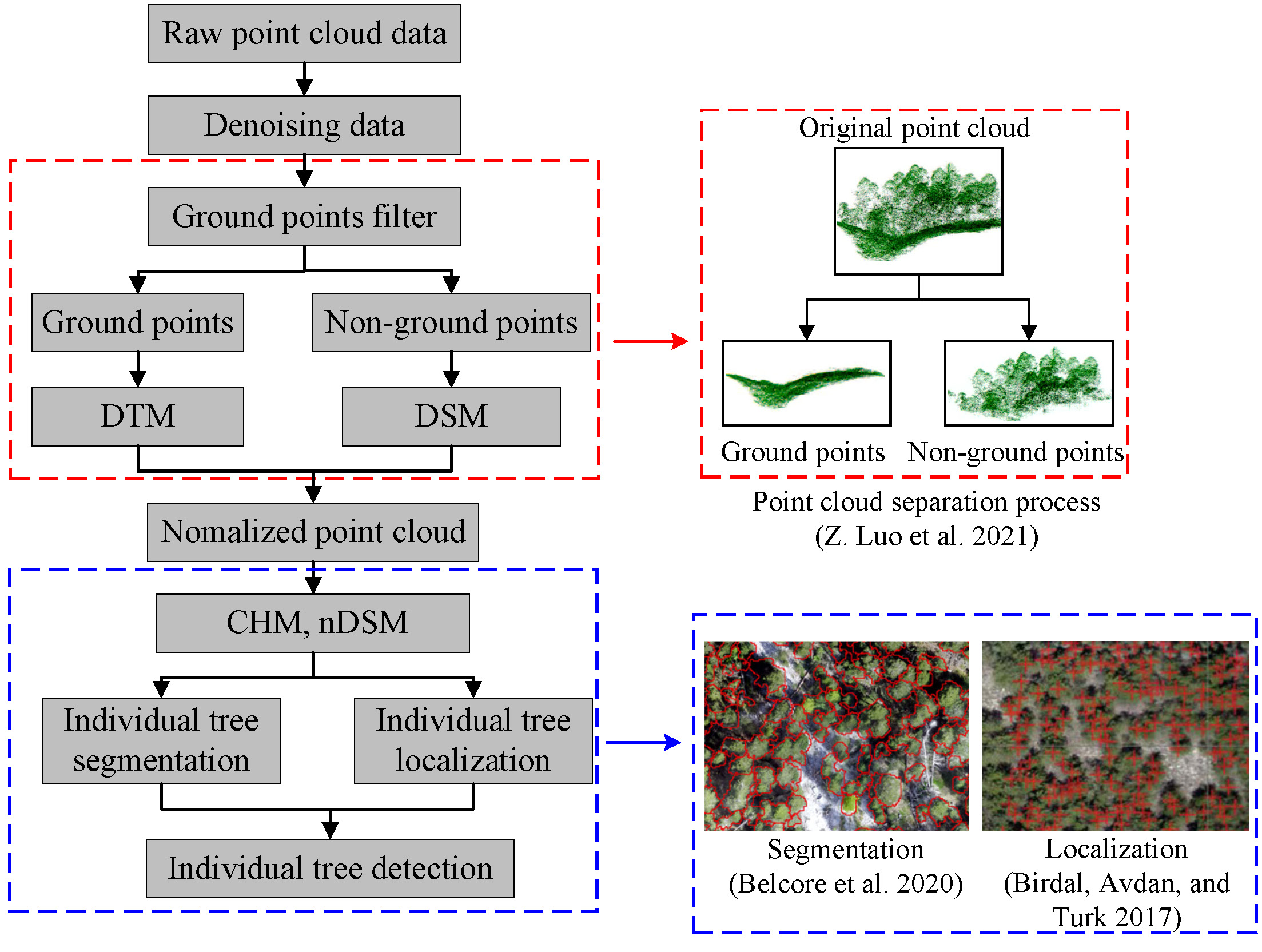}
  \caption{Typical study workflow of the tree detection based on 3D point cloud \parencite{birdal2017estimating,belcore2020individual,luo2021detection}}
  \label{fig:Workflow1}
\end{figure}

\begin{figure}[ht]
\centering
  \includegraphics[height=12cm]{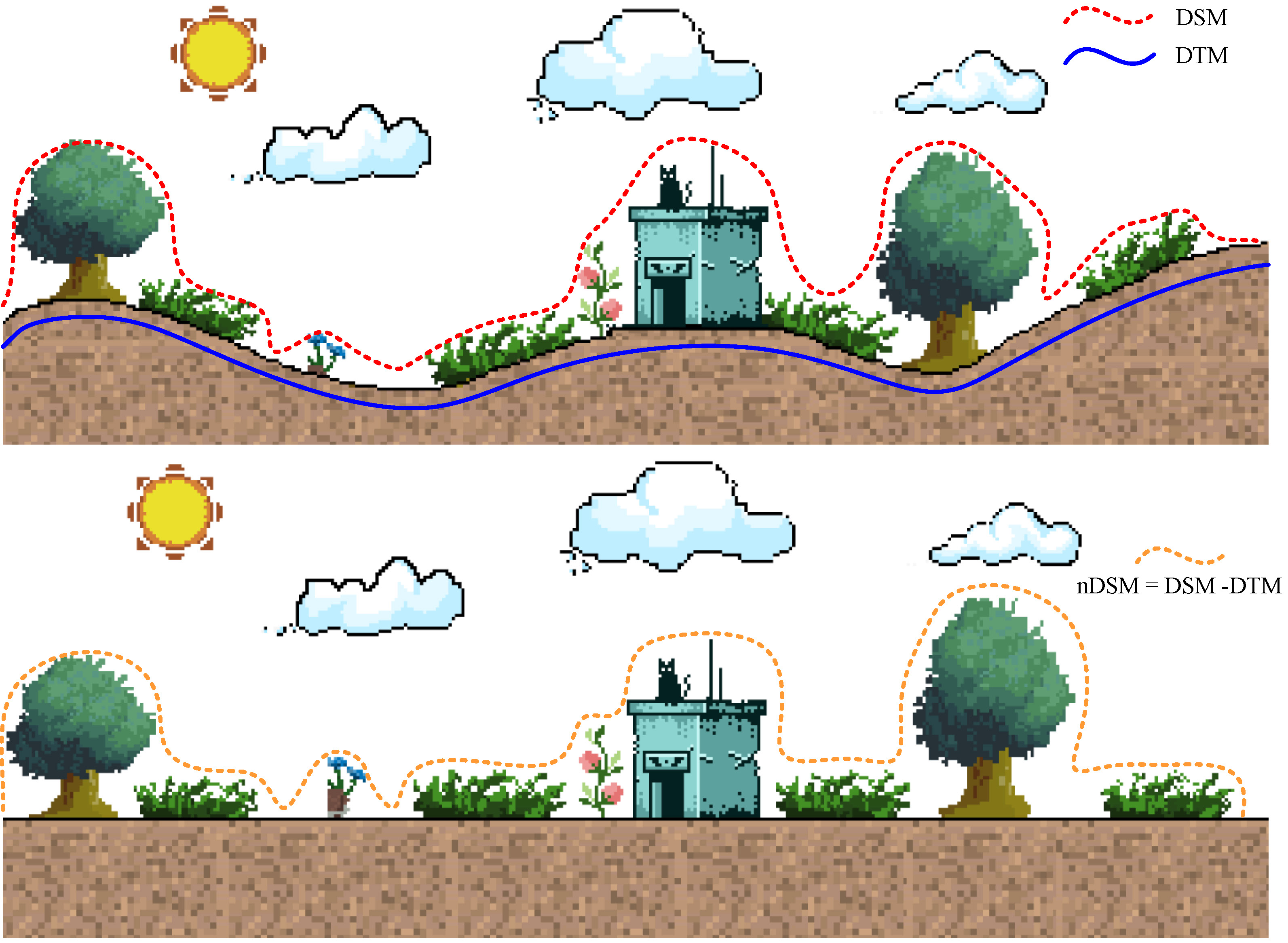}
  \caption{Schematic illustration of DSM, DTM, and CHM}
  \label{fig:CHM}
\end{figure}

The 3D point cloud data has been commonly used on the ITD in remote sensing. The main ways to obtain 3D point cloud data of the plantation are through LiDAR and DAP. The former obtains three-dimensional information about the forest directly by transmitting and receiving laser pulses using active sensors, Whereas DAP extracts the 3D point cloud data by processing aerial images using passive sensors. Fig. \ref{fig:Workflow1} is the typical study workflow of tree detection using 3D point clouds. In the process, the raw point cloud data will be pre-processed to remove the noise at first and then separated into ground points and non-ground points by a filter. The non-ground points will further be interpreted to generate different models, e.g., Digital Terrain Model (DTM), Digital Surface Model (DSM), and normalized Digital Surface Model (nDSM). Fig. \ref{fig:CHM} illustrated the DTM, DSM, and nDSM schematically. The DTM is depicted by the blue curve which is the bare ground surface with removing all other features. The DSM is represented by the red curve which includes both the natural and artificial features. And the orange curve shows the nDSM which represents the absolute height of all objects after removing the ground point, is characterized by subtracting DTM from DSM. When nDSM is applied in forest areas, nDSM is also called the Canopy Height Model (CHM) because most objects are trees.
Subsequently, individual tree segmentation and localization will be conducted based on these models. In the end, the ITD is achieved by combining the results from segmentation and localization
 
\subsection{Point cloud data collection and pre-processing}
The forest canopy structure could be reconstructed in forestry management by high-quality point cloud data. Classifying by different platforms, LiDAR-based point cloud collection methods can be divided into different kinds of LiDAR scanning, e.g., spaceborne, airborne, UAV, and terrestrial scanning. This review focuses on UAV LiDAR scanning that mounts a sensor on a UAV and acquires data rapidly and conveniently in a top-down approach. In the data acquisition process, the main parameters are the flight altitude, flight belt interval, flight speed, emission frequency of the laser pulse, scanning angle, and overlap. For the result of point cloud data, the key parameter is point cloud density. Typically, the point cloud density is between 100 and 300 $pts/m^2$. Researchers sometimes use a very high overlap rate of 90 \% for acquiring high point densities of more than 1000 $pts/m^2$.

On the other hand, point cloud data can also be acquired from UAV images by SfM techniques \parencite{ghanbari2021individual,gonroudobou2022treetop,klouvcek2022uav}. In the flight plans, flight altitude, flight speed, overlap rate, camera parameters, and ground control points play the important role in acquiring UAV images. Among them, ground control points are points on the surface as known locations and used for geo-reference data \parencite{ yancho2019fine,donmez2021computer}. Moreover, most SfM processes are completed by commercial software, such as Agisoft Metashape software \parencite{belcore2020individual,polat2020investigation} and Pix4Dmapper software \parencite{johansen2018using,huang2019leaf,sun2019remote}. 

Due to the multifarious types of the terrain and plant, the original point cloud data are complex and diverse. Therefore, the data used for the ITD should be normalized at heights and only includes trees. The first step of data process is to filter the noise of the  data. Most researches use the commercial software for data processing, especially denoising data, such as Lidar 360 software \parencite{hu2020development,ma2021novel,ma2022performance}, rLiDAR and LiDAR packages in software R \parencite{roussel2020lidr,da2021using}, FUSION/LDV software \parencite{mcgaughey2012fusion,guerra2018comparison,moe2020comparing}, CloudCompare software \parencite{hadas2019apple,ghanbari2021individual}, LASTools software \parencite{conrad2015system,yancho2019fine,huang2019leaf} and Terrasolid software \parencite{puliti2020estimation}. 

After reducing the noise of the data, a ground point filter will separate the ground points and non-ground points, which can be further used to generate DTM and DSM. The mainstream approaches for ground point filters are as follows:

As developed in year 2016 \parencite{zhang2016easy}, Cloth Simulation Filtering (CSF) has now become one of the most commonly used ground-filtering algorithms with few parameters to be set, high accuracy  \parencite{liu2021predicting,rudge2021modelling,lei2022novel}. Before introducing CSF, the meaning of cloth simulation should be known first. Based on Mass-Spring Model \parencite{provot1995deformation}, cloth simulation can build the cloth with many interconnected particles with mass. The location of the particles determines the position and shape of the cloth. The interconnection between particles is a virtual spring which obeys Hooke's law and has three types, such as traction spring, shear spring, and flexion spring. The relationship between position and forces on particles could be calculated using Newton's second law of motion.

In general, the CSF used as a ground-filtering algorithm could be summarized in four steps: (1) put a cloth on the inverted surface, which is transformed from turning upside down the point cloud. This cloth includes particles and interconnections; (2) under the force of gravity, particles drop along the terrain, and the displacement can be calculated; (3) particles intersect with the terrain will be set as unmovable ground points, and the position of particles can be modified based on the internal forces; (4) the final shape of the cloth can separate ground points and non-ground points and further build a DTM. Additionally, some conditions are used to simplify the model and make it feasible. The particle is constrained to move only in the vertical direction, so the collision can be detected by comparing the height values. The mass of particles and rigidness are both constants. However, the cloth will make significant errors at the place with the sparse point cloud. In order to overcome this shortcoming, a modified CSF algorithm make the rigidness of each particle adaptively changes with the point density of the initially bare earth point cloud \parencite{lin2021comparative}. Even using the sparse point cloud, the improved adaptive CSF algorithm still shows a good performance to generate the high quality DTM. 

The second ground point filter is the progressive Triangular Irregular Networks (TIN) filtering algorithm \parencite{axelsson2000generation}. In this algorithm, original data is filtered for only keeping the measurements belonging to the ground surface. Firstly, using all collected point cloud data, an initial sparse TIN is generated from the original lowest seed points. Then, the TIN adapts the data iteratively and is constrained under the curvature set in advance in the model. Points can be added to the TIN if the value is below threshold values, such as distance and angle. Finally, the TIN is gradually densified in the iterative process, and all points are classified as the ground or object. Because the progressive TIN filtering algorithm is robust in processing the random noise point cloud, this algorithm is suitable for filtering the 3D UAV point cloud \parencite{zhang2018filtering,zeybek2019point}. However, the traditional TIN filtering algorithm is usually applied on specific sites, such as flat and low-vegetated regions. Therefore, an Improved Progressive TIN Densification (IPTD) filtering algorithm was developed and performed well in various complex forested landscapes \parencite{zhao2016improved}. The IPTD filtering algorithm has been used in many researches \parencite{wang2019improved,zhou2022fusion}. In this improved algorithm, instead of choosing the lowest points, a morphological opening operation is used to acquire more ground seed points that are evenly distributed. To avoid the generation of redundant triangles, it also adds the simulated ground points to the original TIN to improve the quality of the TIN.


\subsection{Individual tree detection}
Since the tree point cloud is distinct from other tiny object point clouds in many scenarios, individual tree localization and segmentation can be seen as a kind of individual tree detection. This review presents some standard algorithms used in the ITD. Generally, the detection of trees requires the combination of several algorithms which play roles of segmentation and localization during the process.

LM algorithm is used as a localization algorithm based on the hypothesis that the highest value is on the tree apex in CHM \parencite{da2021using,ma2022performance}. A window with given size is set to traverse the CHM, generating a local maximum point for each move. If the local maximum is higher than the threshold representing the minimum height of a tree, this local maximum point is taken as a treetop. This process means that one tree is detected. In this algorithm, CHM resolution and window size play essential roles in tree detection. For example, one tree will be regarded as two separate trees when the window size is too tiny \parencite{yin2019individual} because the big and tall tree will be detected more than two times during the move of the small window \parencite{guerra2018comparison,liu2020dominant}. 

WA is a segmentation algorithm often used with LM to achieve the ITD \parencite{wu2019assessment}. This method is based on simulating immersion and could sense subtle changes and generate closed contour lines around objects \parencite{persson2002detecting}. In the ITD, it can build a barrier at canopy boundaries as watershed boundaries. Such boundaries prevent the merging of adjacent waters. In other words, they can segment the adjacent tree crowns. In WA, the core parameter is the threshold value of minimum height and sigma which influence on the number of segmented trees directly. 


 However, WA does not perform well in the forest area with higher tree densities because the segmentation line of the overlapping tree cannot accurately represent the actual canopy area. After that, a progressive watershed segmentation algorithm is proposed to improve the segmentation performance in such situations as identifying the overlapped crown areas \parencite{ma2021novel}. Identifying overlapped areas is based on observing the relationship between the point cloud density and the enclosed irregular curve. The theoretical radius of the curve and the derivative of the radius increment are essential parameters during the identification. For instance, one area can be considered as a single independent tree when the derivative of the increment of the theoretical radius is equal to zero.

\begin{figure}[ht]
\centering
  \includegraphics[height=7cm]{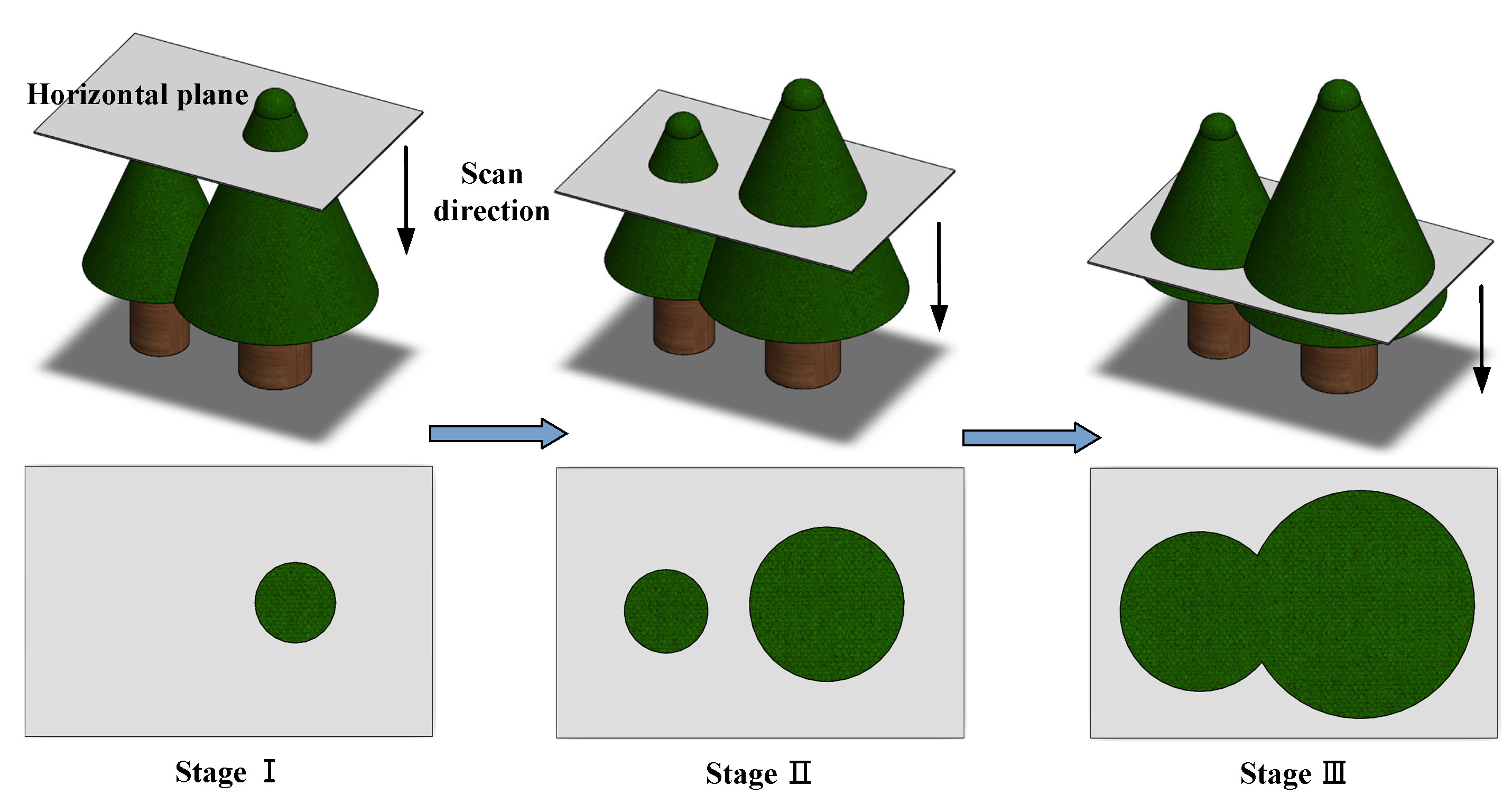}
  \caption{Schematic diagram of RHCSA algorithm \parencite{zhao2017region}}
  \label{fig:RHCSA}
\end{figure}

Usually, the ITD includes two steps as tree top detection and tree crown segmentation. A one-step algorithm called Region-based Hierarchical Cross-section Analysis (RHCSA) is proposed to complete the detection and segmentation together \parencite{zhao2017region}. The algorithm treats the tree crown as a 3D topographic surface, which can be represented as a mountain-like uplift on the CHM. In this approach, the CHM is scanned vertically by a horizontal plane within the specified interval, as shown in Fig. \ref{fig:RHCSA}. During the scanning process, when the plane passes through trees, the cross-section areas will be created from the higher tree to the lower tree in sequence. After the scan, the CHM is decomposed into a horizontal crown area of different heights. This algorithm will automatically determine whether the cross-section appears first. When different individual tree crowns do not have contact with each other, the cross sections show multiple circulars; when they interact with each other, the cross sections usually show an irregular shape. Users can define the specified interval as starting and ending heights.

Moreover, the iteration time could be calculated with the height of the level cutting in a single scan. The scanning accuracy could be improved by increasing the step subdivision. In the end, the output of this algorithm is delineated as individual tree crowns after the iteration. However, this algorithm is restricted by the specified interval that it cannot delineate the small trees which are outside the starting height and ending point \parencite{liu2021predicting,rodriguez2021uav}.

 Point Cloud-Based Cluster Segmentation (PCS) is an algorithm that uses a spacing threshold, a minimum spacing rule, and a horizontal profile of the tree profile \parencite{li2012new}. This algorithm classifies points by comparing the distance between the target point and other treetop points. The distance greater than a specified threshold from the treetop points will also be excluded. In order to solve the problem in classifying the lower level points, which may involve the overlapping area, points are classified from the highest and lowest sequentially. Moreover, it can also add a shape index to reduce the error when classifying whether points belong to an elongated branch or a separate treetop. One of the shortcomings of PCS is that it is highly dependent on the value of the spacing threshold. Therefore, the PCS does not perform well in a complex forest with many overlapping trees \parencite{hu2020development,lu2020estimation}.

A forest tree segmentation algorithm combined k-means clustering, PCS, and LM was proposed as named Layer Stacking (LS) \parencite{ayrey2017layer}. In the LS, all objects are layered at one-meter intervals horizontally through the full height. LM is used to find the maximum height point in CHM representing each tree top. And then, every layer is applied with k-means clustering using maximmu points as seed points for clustering points. After clustering, each cluster is placed with a polygonal buffer around it. The purposes of the buffer are to be an additional round of clustering to reduce the classification errors and to be a means to connect the points and vectorize the clusters. A rasterized map of the overlapping polygons is then generated. Next to that, the overlap map is smoothed to remove different overlapping areas in the tree that might represent branches. In the end, LM is used again on the smoothed overlapping map to find the centers of trees. Each local maximum is placed with the buffer and takes its centroid as the new center point for the tree. This step helps in merging local maxima to prevent them, which are too close to each other, from being classified into separate trees. The last local maxima are regarded as the centers of each tree and will be used for tree segmentation later.

A study compared the performance of Watershed, PCS, and LS in three tree datasets, i.e., leaf-on, leaf-off, and fused (combine leaf-on and leaf-off) point cloud \parencite{chen2022individual}. In this research, three individual tree segmentation algorithms were applied to the four forest types with three datasets. As shown in the results, WA, PCS, and LS performed similarly in leaf-on and leaf-off datasets. Regarding fused point cloud data, the LS algorithm had the higher F-score than that of the WA and PCS algorithms. For example, the segmentation had 15.7 \% more trees with the fused data than leaf-off data in the mixed broadleaved forest. The high point density caused by fused point cloud data includes more small trees and is responsible for the improved segmentation.

Another research analyzed the sensitivity of critical parameters in WA, LM, PCS, and LS for choosing a superior method in different situations \parencite{ma2022performance}. The WA and LM methods were implemented based on the data from the CHM when the PCS and LS were based on the Normalized Point Cloud (NPC). Results indicated that the NPC-based algorithm showed a better capability than the CHM-based algorithm in tree segmentation. For instance, the average detection rate r of the NPC-based method is 87.06 \% which is higher than that of the CHM-based algorithm with 81.28 \%. The limited pixel resolution of CHM rasters causes a lower detection rate.

On the one hand, the core parameters, such as the sigma value in WA, the window size in LM, and the distance threshold in PCS, showed a negative correlation with the detection rate performance. On the other hand, the layer thickness in LS was positively correlated with the detection rate. Meanwhile, the results also showed that the WA and LM did not have the advantage on the tree segmentation with a high-density point cloud. The PCS algorithm was good at segmenting many types of forests. Furthermore, the LS algorithm achieved better results in multilayer and extremely complex heterogeneous forests.

Some methods perform on basis of point density without relying on the tree height \parencite{yang2019influence}. For example, Adaptive Mean Shift 3D (AMS3D) segmentation  \parencite{ferraz20123} is first used in the point cloud collected from airborne laser scanning data. As a multi-modal distribution, it classifies the point cloud as many models where each model represents an individual crown with the maximum local value in density and height. There are two steps in this approach. In the first step, a 3D kernel is moved to denser regions iteratively until convergence for acquiring multiple modes from the point cloud. In the second step, the points belonging to the same mode are gathered to compute the 3D clusters of the individual tree crown. Because AMS3D is not a parametric algorithm, it does not pre-set the contours of objects in the data and can be used to analyze complex forests with different structures and shapes. The only influential parameter is the size of the 3D kernel named bandwidth. For example, the small bandwidth will divide the giant tree into smaller trees due to the search for the local maxima in a narrow section. The single bandwidth cannot meet the requirements to process the point cloud data with different sizes of the tree crown. Therefore, a study established a self-calibrated bandwidth model which can be optimized with different forest structures \parencite{ferraz2016lidar}. That is to say, the bandwidth increases from the lower points to the higher points \parencite{ rudge2021modelling}. Moreover, the segmentation accuracy of the algorithm could also be improved by using different canopy structure attributes to calibrate the bandwidth adaptively \parencite{lei2022novel}. To verify the generalization of the improved algorithm in different species, it is applied in a subtropical forest with an overall precision greater than 0.72.

The algorithms, which rely on the local maximum height to find the tree top location, do not perform well when dealing with trees without apparent tree tops or irregular tree tops. As a very robust method, the Random Sample Consensus (RANSAC) algorithm can address such problems \parencite{balsi2018single}. Initially, Roth and Levine \parencite{fischler1981random} proposed RANSAC to extract geometric primitives in model-based computer vision. This algorithm searches the model by comparing the point cloud with a specified minimal set of points, the smallest number of points needed to define a geometric primitive. The remaining points are used to fit the shape after the minimum number of points has been selected\parencite{polat2020investigation}. Because RANSAC carries out the whole point cloud data, extracting all the shapes similar to the parametric model, the candidate Region of Interest (RoI) is needed to process only the most relevant point cloud to increase its efficiency and reduce its computational cost. 

Another common problem in tree detection is that many algorithms detect the tree based on a top-down structure. For such algorithms, the tree top is localized followed by segmenting the tree crown. And then, the tree trunk can be delineated in the end. However, it is difficult for this approach to perform well in forest with many interlacing tree branches. Since it takes much work to complete the tree top localization or tree crown segmentation, some researchers design their algorithm based on the bottom-up scheme \parencite{lu2014bottom,lin2021leaf}. The basic flow of the bottom-up algorithm is to first locate and delineate the trunk, and the tree crown can then be delineated by clustering the points around the trunk.
Moreover, the core concept of trunk localization is that the trunk corresponds with higher point density and elevation \parencite{bayat2019individual}. Subsequently, the tree crown segmentation is conducted based on the fact that points belonging to its trunk have a smaller distance than any other trunk. This concept is similar to the PCS algorithm. Since the distance between the points and the trunks is a core parameter, the algorithm still needs to be improved when a branch reaches into the crown of another tree.

Nowadays, more and more researchers conduct tree detection based on machine learning, especially DL. For example, based on CHM, a DL framework is used to detect trees in the forest with complex structures with mining defining features \parencite{luo2021detection}. There are three steps in the tree detection stage: Firstly, to obtain the vertical structure information in a different layer, the point cloud is sliced into several parts in a top-down way; Secondly, three properties, i.e., height, height gradient, point density, as a three-channel representation are inputted to the DL network simultaneously for providing valuable information. The height gradient could show gaps between the adjacent tree crown. The point density helps extracting the tree because the center of the tree has a higher point density. Thirdly, a multibranch network is built to extract trees from grid images. This research uses region growing algorithm and Mask RCNN to compare the performance with the proposed multibranch network method. Results show that the new method performs well than the other two methods achieving a mean precision of 85.99\%, a mean recall of 89.23\%, and a mean F1-score of 87.04\%.

\subsection{Comparison and combination of tree detection between LiDAR and DAP}

Both the point cloud data acquired from the DAP and LiDAR work well in tree detection. Therefore, some researches explored to compare the differences between them and combine them to find superior detection performance in different scenarios.

There are some studies focused on combining DAP and LiDAR technology for improving the performance in extracting the individual tree crown information. For example, in a multi-resolution segmentation algorithm, researchers combined the RGB information from photogrammetric images and CHM from LiDAR-normalized point clouds for tree crown segmentation. In this algorithm, the weight of the CHM layer is three times higher than that of the RGB layer \parencite{moe2020application}. However, results show a lower correlation between field-measured and multi-resolution segmented crown areas for all species. The complex crown structures in deciduous species and the varied spectra may cause inaccurate individual tree crown segmentation. Many previous studies also proved that it was difficult for tree crown detection algorithms to perform well when the object was broadleaf species \parencite{vauhkonen2012comparative,wang2016international}.

Another study combined the DAP-based DSM and the LiDAR-based DTM \parencite{klouvcek2022uav}. Although such a combination increases the height accuracy of the normalized surface model, the accuracy of tree crowns height and detection do not improve. Therefore, the researchers conclude that DAP could replace LiDAR system when the duty performs a task with frequently changing sites to save costs. Furthermore, combining two types of data lead to the better quality of the DTM.

On the other hand, one research compared two data acquisition methods, i.e., DAP and LiDAR on the difference of the point cloud quality and the estimated tree crown parameters over an urban park \parencite{ghanbari2021individual}. The result shows that the types of point clouds are highly correlated with the statistical measure of fit R$^2$ up to 99.54 \%. Moreover, the estimated tree height, diameter, area, and volume results correlate that R$^2$ is higher than 95 \%. 

It is not feasible to compare the points directly because of the differentiation between the center positions of two corresponding LiDAR and DAP points. So, a study compares the quadratic rasterized points and CHMs generated from LiDAR and DAP points. In the raster, the highest height of points is set as the value within each cell \parencite{thiel2017comparison}. Results indicate a high correlation rate between LiDAR and DAP data. Compared to the LiDAR data, the DAP data has more detailed information in geometry which causes the advantage in tree detection. In the same area, with the same algorithm, the detection rate using DAP data is 93.2 \% which is higher than that of LiDAR data at 78 \%. There are two main reasons for this phenomenon. Firstly, the DAP data has a higher point density which includes more detailed information about the plantation. Secondly, the penetration of the LiDAR pulses is slight, which ignores the small targets with lower reflection.

\section{Object Detection based on Spectral Channel in Images}
The methods used in the ITD with spectral images could be categorized into the DL-based and classical methods. Most DL methods applied in detection are based on the object, such as the algorithm of the YOLO series, R-CNN series, and SSD, which mark the target with boxes on the image. On the other hand, The classical methods detect trees based on the pixel and classify every pixel into different classes. 

\subsection{Spectral image data acquisition}
The spectral image could be roughly divided into RGB images and multispectral images, which are collected by RGB, e.g., senseFly.S.O.D.A, and multispectral, e.g., Parrot Sequoia and  RedEdge-M, cameras, respectively \parencite{osco2020convolutional,jintasuttisak2022deep}. In general, the acquisition location, weather, and the model and parameters of the UAVs should be clearly described  \parencite{sun2022detection}. The flight mission, altitude, speed, and spatial resolution are the main parameters in acquiring spectral images \parencite{lou2022measuring, zhang2022multi}. The spectral images used in DL methods are divided into training, validating, and testing datasets. In the training and validation datasets, all the images should be labeled with bounding boxes \parencite{sun2022detection}. In order to decrease the influence of weather, equipment, and suspended particles on the quality of images, a balance contrast enhancement technique could be used \parencite{safonova2022detection}. In this technique, the histogram pattern of the input image remains the same, with the contrast that can be stretched and compressed \parencite{guo1991balance}. 

\subsection{Individual tree detection}
\begin{figure}[ht]
\centering
  \includegraphics[height=10cm]{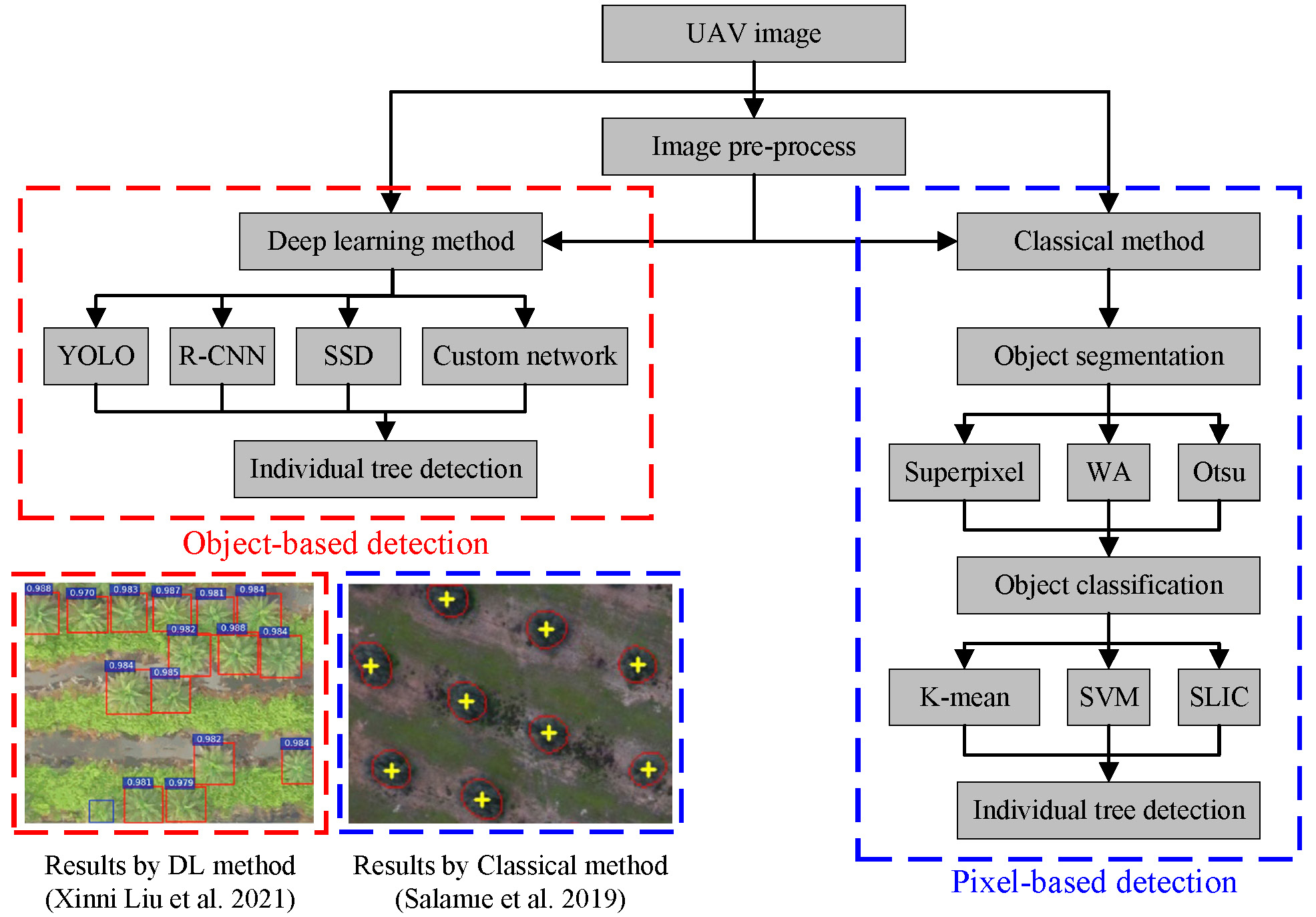}
  \caption{Typical study workflow of tree detection based on spectral images \parencite{ salami2019fly,liu2021automatic}}
  \label{fig:RGB images detection}
\end{figure}

Fig. \ref{fig:RGB images detection} shows the workflow of the tree detection with spectral images. As shown in the workflow, the classical method includes object segmentation, e.g., superpixel and WA, and object classification, e.g., k-mean, mean-shift and SVM. In addition, there are still extraordinary examples. For instance, in the DL methods, a study used the ResNet-18 network for semantic segmentation by classifying pixels into different palm trees, i.e., a pixel-based method, \parencite{ferreira2020individual}. Moreover, in the classical methods, as an object-based method, the Template Matching (TM) method finds targets by comparing the template with objects in the image \parencite{avtar2020unmanned}. 
Since the pixel-based detection method classifies every pixel singly, the variations in spectral patterns from changing weather and season impact the accuracy of the method performance significantly\parencite{egli2020cnn}.
Tree detection results by DL and classical methods are also shown in Fig. \ref{fig:RGB images detection}. Generally, the detection results acquired by the DL method usually used the boxes to mark the trees, while the results by the classical method marked the trees with specific contours.

\subsubsection{Detection with DL methods}
The DL methods have been widely used in the object detection duty due to the technological advances in graphics processing unit. Generally, DL-based tree detection is supervised approaches and need a large training set to learn. The quality of the training set is crucial to the performance of the detection. Among them, Yolo series algorithms and RCNN series algorithms are applied on the tree detection popularly. Other than that, some detection methods will process the image before applying the DL methods to achieve a better detection result. Commonly, the processing methods are noise removal and potential areas selection.

\paragraph{YOLO series algorithms}

As one of the one-stage object detection methods with the highest performance, the YOLO series algorithms take a single forward pass on the input images and outputs the detection results. In the results, the detected targets are marked with the bounding box and class probabilities \parencite{lee2023air, wu2020using}. Due to the excellent performance, the YOLO series algorithms have been applied in detecting plantation widely \parencite{mirhaji2021fruit, choi2022automatic, ho2022vision}. 
For example, one research used YOLO-V5 to detect palm trees from RGB images and compared it with YOLO-V3, YOLO-V4, and SSD300 for evaluating their performance \parencite{jintasuttisak2022deep}. In particular, YOLO-V5 has four sub-versions as  V5s (small), V5m (medium), V5l (large), and V5x (extra-large), which have the same network structure but with different depths from small to large. The results showed that the YOLO-V5m had the maximum mAP values of 92.34 \%. Furthermore, the YOLO-V5 method had the shortest average detection time compared with YOLO-V3, YOLO-V4, and SSD300.

In order to explore the superiority of the YOLO algorithms, many researchers conducted experiments to compare tree detection performance between YOLO with other deep leanring networks. For instance, one study compared the performance of Faster R-CNN, YOLO-V3, and SSD on the detection of loblolly pine crowns \parencite{lou2022measuring}. Moreover, another investigation applied the SSD, YOLO-v3, YOLO-v4, and Fast-RCNN in a spruce forest area to compare their performance on the tree detection \parencite{emin2021target}. Meanwhile, in another research of the assessment of CNN-based methods for the ITD, Faster R-CNN, YOLO-V3 and RetinaNet were compared by being applied on the UAV images \parencite{santos2019assessment}. RetinaNet obtained the most accurate results with an average precision of 92.53 \%. YOLO-V3 achieved the fastest computation speed compared to its counterpart. Moreover, Faster R-CNN has the lowest accuracy and highest computational cost.

Although the YOLO algorithms achieves good accuracy in tree detection \parencite{safonova2022detection}, some structures can still be improved and optimized. For example, even if YOLO-V4 can extract the feature effectively, the complex backbone network, i.e., CSPDarknet53, in YOLO-V4 increases the detection time. Therefore, one study improved the YOLO-v4 in three steps \parencite{sun2022detection}. Firstly, the CSPDarknet53 was replaced by MobileNetv2 as a backbone network to increase efficiency while maintaining its accuracy. Secondly, Convolutional Block Attention Module (CBAM) was added to ignore the redundant features, prioritize the essential features, and further optimize the detection processes. Thirdly, for deepening the network depth and reducing the complexity of the calculation, the original 3 $\times$ 3 convolution kernel for information integration on the YOLO-V4 was replaced with 1 $\times$ 3 and 3 $\times$ 1 convolution kernels. Subsequently, this improved DL network was used to detect pine wilt nematode. The results showed that the improved algorithm had a better performance than its counterparts, such as Faster R-CNN, SSD, and YOLO-V4 regarding average precision, training time, parameter size, and test time.

The YOLO series algorithms commonly concentrate on detecting trees with large sizes and outstanding features. For the detection with a small size and few features, one research proposed a lightweight, small object detection method based on the YOLO network toward the detection of dead trees on mobile terminals \parencite{wang2022lds}. The proposed LDS-YOLO algorithm made improvements in the following aspects. At first, to resolve the problem of overfitting caused by the small-scale dataset, the dense connection module was added in the backbone part. By combining different feature layers in dense connection modules, the high-dimensional feature reused the low-dimensional features to improve the regularization and smooth decision boundaries. Second, a variant of pooling  SoftPool \parencite{stergiou2021refining} was inserted into the network to minimize the pooling loss and computation cost. In the end, depth-wise separable convolution was used in the network to reduce the number of parameters and further improve the efficiency of the detection. As a result, the number of parameters was decreased by using 1 $\times$ 1 $\times$ 3 convolution kernel.

\paragraph{R-CNN series algorithms}

As a two-stage detection algorithm, R-CNN series algorithms with consisting two parts have been applied to detect trees in many studies \parencite{liu2021automatic}. The first part aims to find the proposed regions on the images, and the second part is then used to process the proposal regions for scoring. For example, Faster R-CNN was used to recognize and extract pine crown \parencite{lou2022measuring} and to extract the spruce tree locations, numbers, and diameters of the spruce tree \parencite{emin2021target}. 

However, for specific detection problems, Faster R-CNN could still be optimized. To detect and observe the oil palm trees, a Multi-class Oil Palm Detection (MOPAD) approach was formulated based on the Multi-level Region Proposal Network (RPN), which plays a vital role in Faster R-CNN \parencite{zheng2021growing}. Based on Faster R-CNN, MOPAD added a Refined Pyramid Feature (RPF) module and a hybrid class-balanced loss module to the detecting process. Benefiting from integrating the multi-level features in the RPF, the network could classify similar classes and detect small objects. The proposed MOPAD method was evaluated using oil palm plantation images and compared to the original Faster R-CNN algorithm. The results indicated that the new method improved the average F1-score by 17.09 \%.

\begin{figure}[ht]
\centering
  \includegraphics[height=7cm]{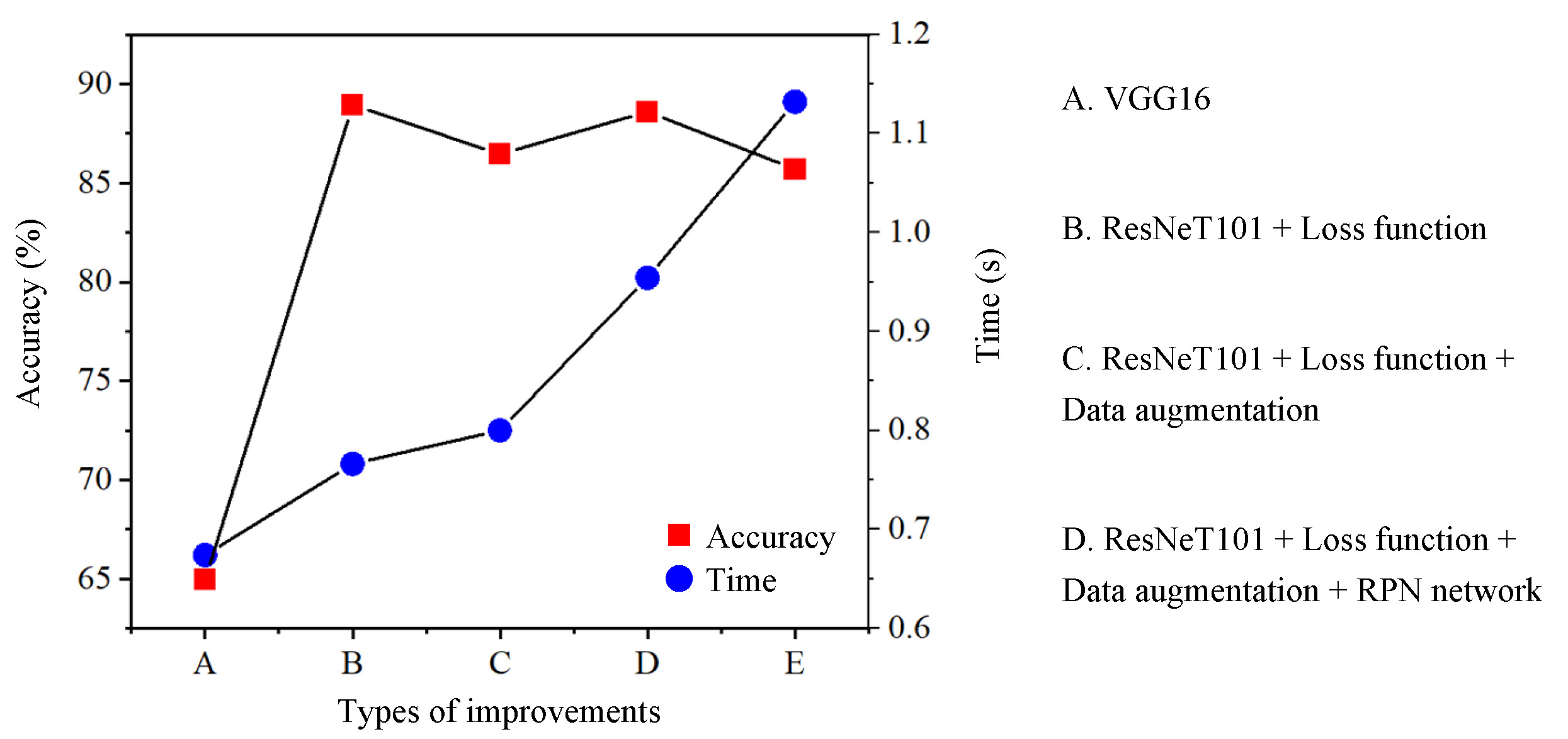}
  \caption{Comparison of the performance in terms of the accuracy and processing time of different improved models \parencite{deng2020detection}}
  \label{fig:Comparison of model performance}
\end{figure}

In another example, a study inherited from the Faster R-CNN algorithm focused on the detection and localization of dead pine wilt trees \parencite{deng2020detection}. This research improved the Faster R-CNN algorithm in three ways. The first one, because the goal of the algorithm was only to detect dead trees as the binary classification problem. The loss function was changed from the Softmax function to the sigmoid function to add the accuracy of the loss value. The second one, the anchor size was optimized according to the small size of detected trees because the anchors were crucial for the pre-selected boxes in the RPN. The last one, the original backbone VGG16 was replaced with the ResNet101 network to deepen the network and prevent the gradient disappearance. As results shown in Fig. \ref{fig:Comparison of model performance}, the detection time increased from 0.649 s to 1.129 s with accuracy increased from 66.2 \% to 70.8 \% by replaced the backbone network with ResNet101. The improved loss function could reduce the calculation time along with the increased time in data augmentation. Meanwhile, the improved RPN network decreased the detection time. In the end, compared with the original Faster R-CNN algorithm, the accuracy of the optimized algorithm was increased from 66.2 \% to 89.1 \%.

Instead of just providing candidate boxes, Mask-RCNN can segment the image to find the same region of the object for achieving image segmentation and classification at the same time in one task\parencite{he2017mask,ocer2020tree, hao2021automated}. 
To delineate trees' properties and observe their growth status, the Mask R-CNN was applied in one study for detecting and mapping the height and the crown width of the individual tree \parencite{șandric2022tree}. Results indicated that the Mask R-CNN showed a good performance in detecting the tree crown with an overall accuracy of over 70 \%. However, in the original Mask R-CNN, the path between the high-level features and low-level features in the FPN for the feature fusion is too long, which leads to the problem in multi-target segmentation \parencite{liu2018path}. 
For improving the algorithm and using it on the large-scale multi-target tree detection, another study modified the fusion style by applying a bottom-up approach which reduced the path of feature fusion between lower and higher layer features \parencite{zhang2022multi}.

\begin{figure}[ht]
\centering
  \includegraphics[height=6cm]{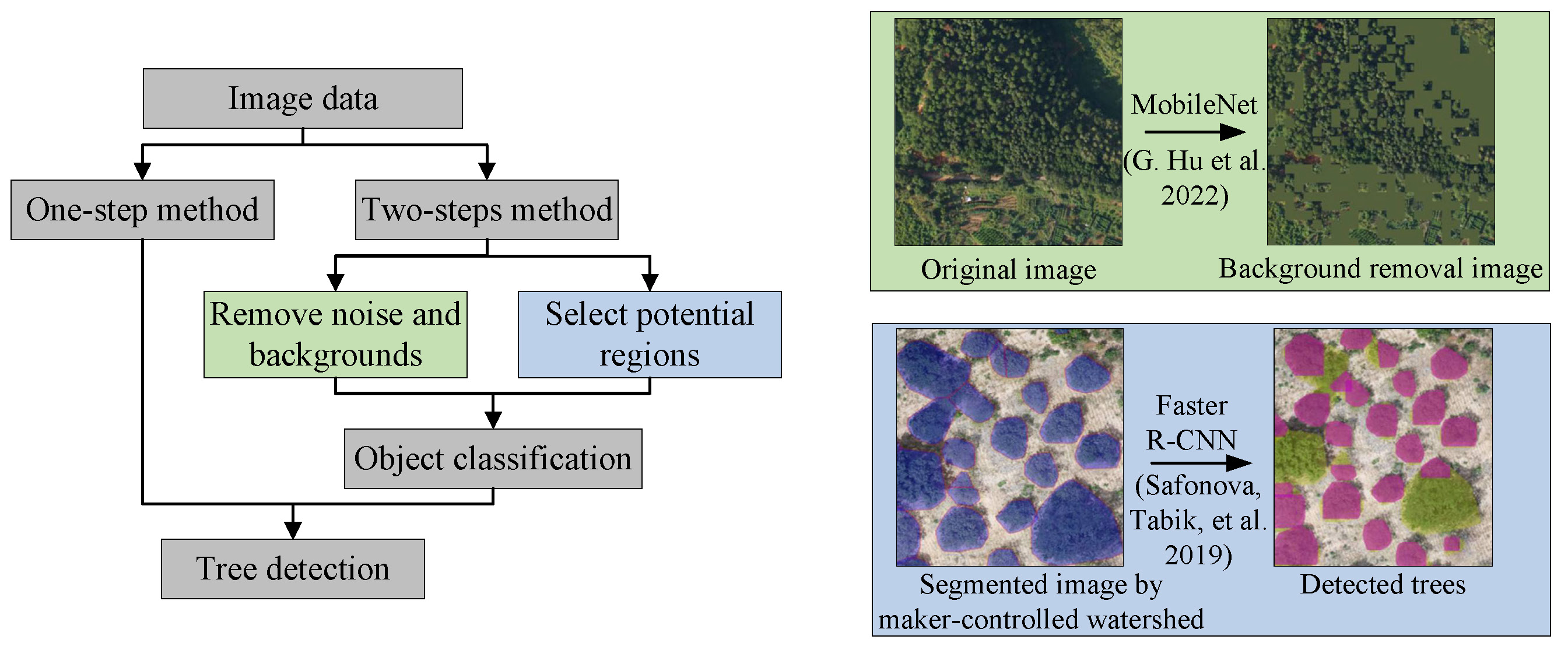}
  \caption{Schematic of the two-steps detection methods \parencite{safonova2019detection,hu2022detection}}
  \label{fig:Two_steps_methods}
\end{figure}

\paragraph{The two-steps methods and other kinds of method}
In addition to the direct application of the DL network on images for tree detection, some methods pre-process the image and then apply the detection algorithm to the result. This two-step method could be classified into two categories, as shown in Fig. \ref{fig:Two_steps_methods}. The first class is to remove noise or backgrounds in advance, while the second class is to select the potential regions before classification.

For instance, in detecting diseased pine trees, MobileNet was first applied to images to reduce interference by removing backgrounds as only interested objects were left in the image \parencite{hu2022detection}. Faster R-CNN was then used to classify the different kinds of trees. This method showed a better performance as higher recall values and lower detection error on tree detection were found, compared to the traditional methods. Moreover, in the research of the detection of diseased pine trees, it used MobileNet to remove backgrounds at first due to the disturbance from other objects, such as shadows and roads. Faster R-CNN was then selected to achieve the detection of disease pine trees.

The second kind of research processed the image first to select the potential regions. The DL network was then used on the selected region to detect the individual tree. For example, in finding the candidate area, the UAV RGB images were transformed into grayscale color palettes at first \parencite{safonova2019detection}. Moreover, the grayscale images were blurred and further applied with a threshold function to create the binary images with thier segmented areas known as the potential regions of trees. After this, the CNN classification model would analyze each potential area. The results showed that the proposed method achieved a better performance than the most famous CNN models in tree detection, such as ResNet, VGG, and DenseNet.

In another example, the results of LM and WA were input to the Faster R-CNN \parencite{malo2021cashew}. The outputs of LM and WA are the information of tree top points and the segmented images as interesting elements. Furthermore, the Faster R-CNN was applied based on the segmented images for tree detection. Compared with the direct application, such as pure classification, this method could increase the efficiency and accuracy of the detection by ignoring the area without trees.
Similarly, in the research on individual sick fir tree identification, some DL methods, i.e., Alexnet, Squeezenet, Vgg, Resnet, and Densenet, classified healthy and sick trees on the images which have been segmented based on LM \parencite{nguyen2021individual}. 

Most DL-based methods conduct object detection based on candidate anchors, meaning that the final prediction results are the score of each box. Some tree detection algorithms do not rely on the candidate anchors but directly make the prediction on the pixels of the image for classification. For instance, to detect and classify the amazonian palms using UAV images, one study used ResNet-18 with DeepLabv3+ on RGB images at first for acquiring score maps with five categories based on each pixel \parencite{ferreira2020individual}. The network can only conduct patch-level semantic segmentation, so it cannot distinguish too close trees. Therefore, the score maps were then performed with morphological operations for the ITD to solve such a problem. The steps of the morphological operations are erosion, opening, dilation, and regional maxima. The results showed that the detection accuracy of the Euterpe precaoria tree and Iriartea deltoidea tree are high, i.e., 98.6 \% and 96.6 \%, respectively. Furthermore, the proposed method can identify two very close trees, which other conventional algorithms cannot achieve.

In another research, a CNN approach is input with RGB images and outputs the predictions of a confidence map \parencite{osco2020convolutional}. In building the training dataset, te confidence map as a dense map is generated in the following ways: (1) Compare the values of every location with each plant location at first, (2) Select the value with the most negligible difference as the value in the confidence map, (3) In the confidence map, the location of the trees could be calculated by using a local maximum algorithm. From the comparison results predicted by proposed approach and other object-based methods, i.e., RetinaNet and Faster R-CNN, the proposed approach showed the highest precision of 0.95 and recall rate of 0.96.

All the methods introduced above can be classified as 2D-CNN, which conducts the tree detection on the RGB images with two-dimensional feature maps. In addition to 2D-CNN algorithms, some researchers used the 3D-CNN on the Hyperspectral Imagery (HI) both in spatial and spectral dimensions. The former usually losses the spectral information in the HI while the latter keeps it \parencite{yu2021three}. Furthermore, most datasets presented so far are RGB images. However, multispectral images are also used in tree detection and delineation, i.e., tree height and crown radius \parencite{yu2021early}. For example, one study applied the Mask R-CNN on the detection using six different band combination data as multi band-DSM, RGB-DSM, NDVI-DSM, Multi band-CHM, and NDVI-CHM combination \parencite{hao2021automated}.

\subsubsection{Detection with classical methods}

\begin{figure}[ht]
\centering
  \includegraphics[height= 8.5cm]{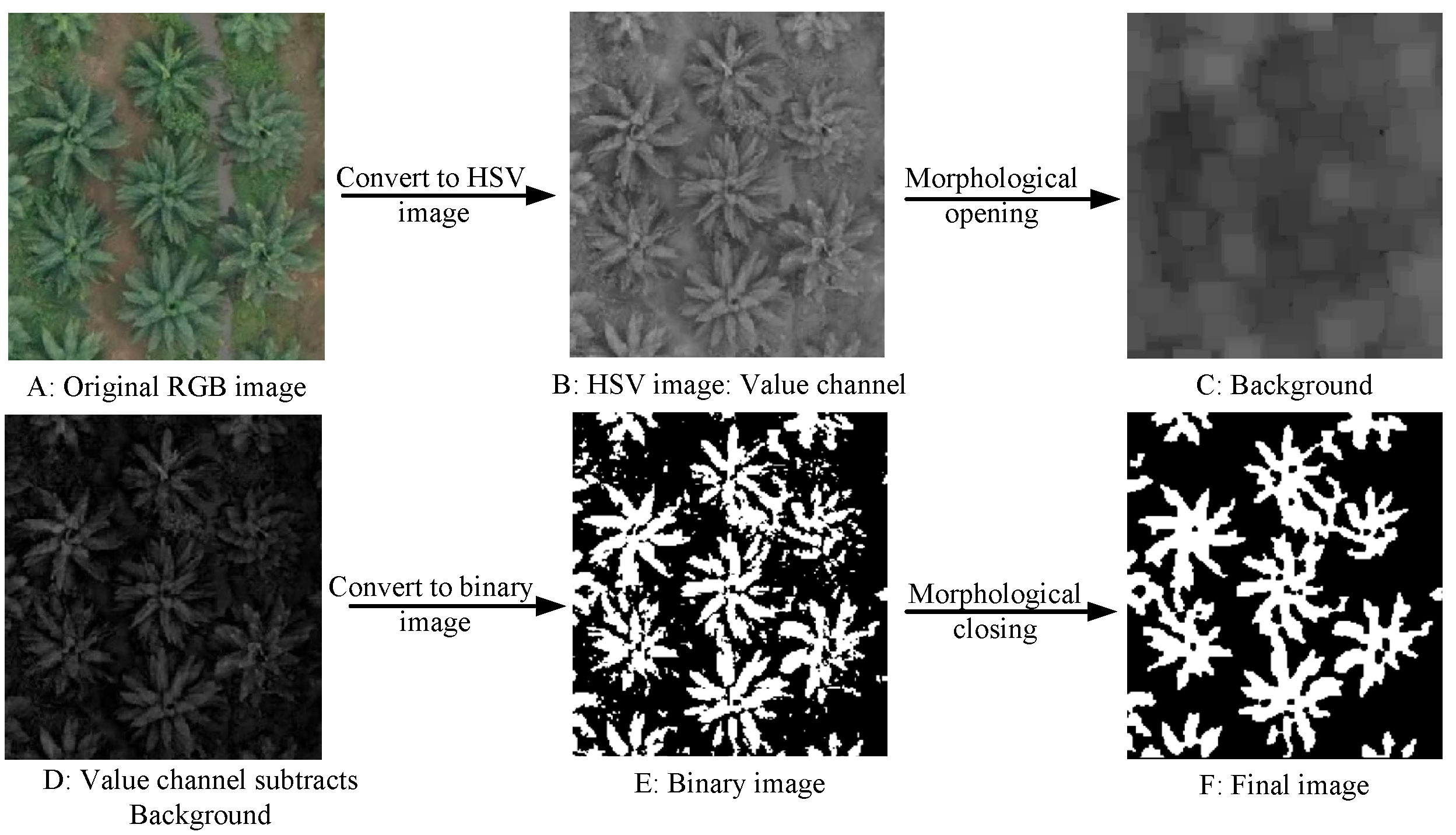}
  \caption{Illustration of the typical morphological operations of the oil palm tree}
  \label{fig:Morphological_operation}
\end{figure}

Most of the classical methods complete the tree detection based on pixel information. In general, the classical methods cannot detect the desired targets directly and alone \parencite{kurihara2022early}. Morphological operation is one of the efficient ways to detect trees initially as it aims to extract interesting structures of the image \parencite{ferreira2020individual, ponce2021methodology}. However, it only can have a good accuracy toward the simple detection due to the single function. Therefore, it is suitable for use in combination with other methods to improve the robust and applicability in the complex detection task. Fig. \ref{fig:Morphological_operation} illustrates the typical morphological operations to get clear binary image of the oil palm trees from the RGB image. Firstly, the RGB image is converted into the HSV image. The background image is obtained by applying a morphological opening operation on the value image from the channel of value in the HSV image. Secondly, the background image is subtracted from the value image. Thirdly, the last image is transformed into the binary image and applied with morphological closing operation. Both opening and closing operations are performed with a square structuring element. Obviously, the main trunk of the oil palm leaves can be extracted well when the background and most of the noise are removed. The original RGB image is acquired from the data supported by an oil palm detection research \parencite{putra2022oil}.

For example, the morphological operation is applied to find the potential regions that have a high possibility to include a tree \parencite{safonova2019detection}. The input images were transformed from RGB to grayscale in the morphological operations. Then, the grayscale images would be blurred by the Gaussian high-pass filter to reduce the noise. Subsequently, the binary images could be created by changing the optimal brightness. In the end, image erosion and dilation could generate candidate boxes of potential regions. And then, the images with candidate regions would be input into a convolutional neural network for classification.

Referring to pixel-based approaches, the superpixel method is used in detection in many ways. For example, one study input the image to a CNN algorithm for getting an initial classification of the trees \parencite{csillik2018identification}. Generally, CNN would cause the single large trees to be detected multiple times, thus treating a tree as multiple. In this research, a classification refinement was based on the superpixel after the initial classification transforms the over-segment imagery into low-level groups. The superpixel was used on the heatmap generated from the results of CNN, and Normalized Difference Vegetation Index (NDVI) layer, i.e., green and bare soil layer using a simple linear iterative clustering algorithm \parencite{achanta2012slic}. After the classification refinement, the detection showed a higher accuracy of 96.24 \%, a precision of 94.59 \%, and a recall of 97.94 \%.

\begin{figure}[ht]
\centering
  \includegraphics[height=7.5cm]{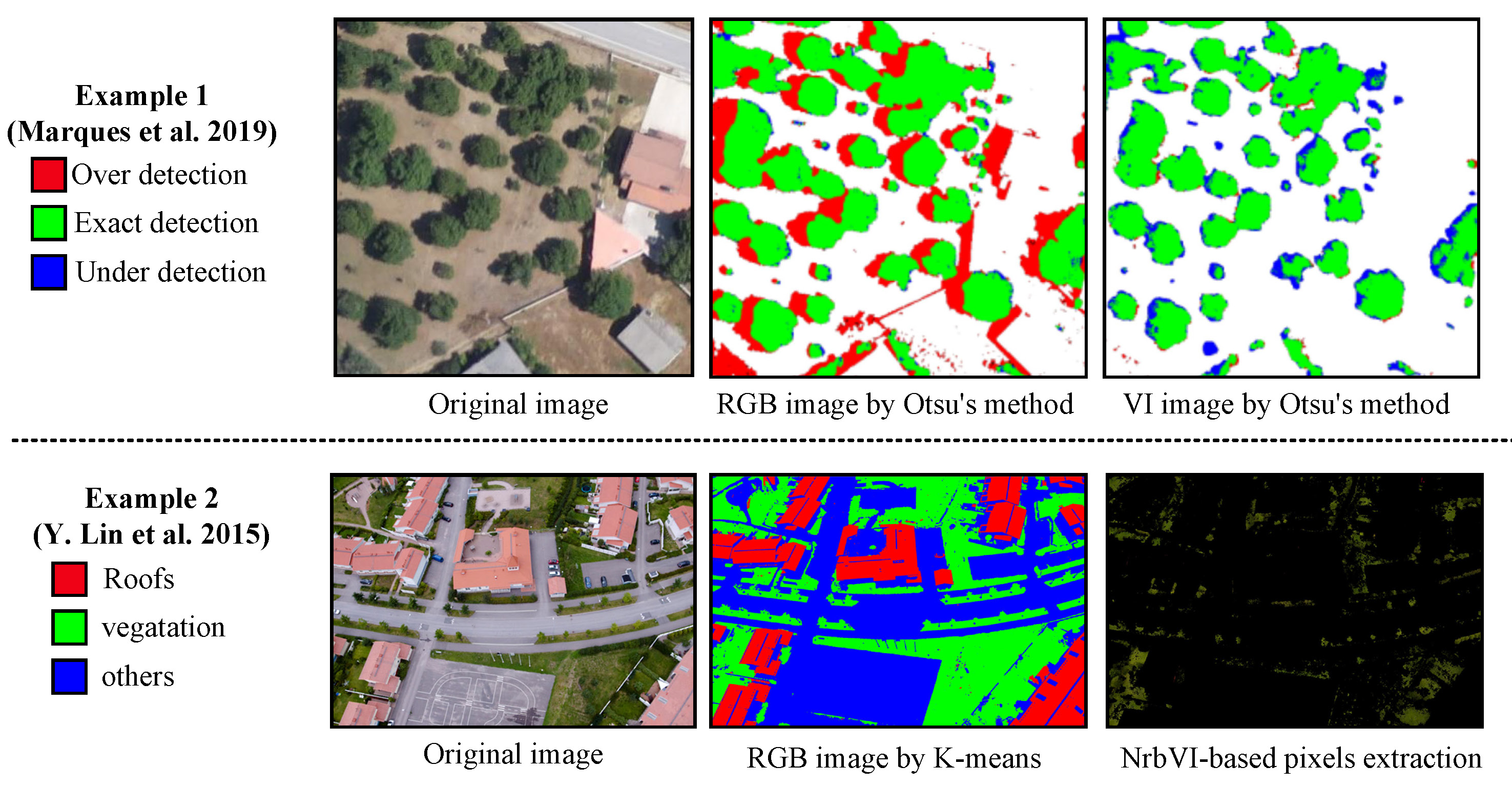}
  \caption{Two examples of image segmentation with and without using vegetation index \parencite{lin2015use, marques2019uav}}
  \label{fig:Comparison_of_VI}
\end{figure}

Moreover, the Vegetation Index (VI) is often used in the classification of plantations as it performs well in distinguishing the vegetation and non-vegetation areas, e.g., builds, roads, and bare soil, in images than the standard segmentation techniques \parencite{garcia2013comparison, lin2015use, marques2019uav,  zhou2022fusion}. Fig. \ref{fig:Comparison_of_VI} shows two examples of image segmentation with and without using VI images. As shown in Example 1, compared with RGB images, a good segmentation result with more exact detection area and the less over-detection area is achieved on VI images. Moreover, in Example 2, the vegetation pixels extraction method based on the R and B bands of the Normalized Difference Vegetation Index (NrbVI) could segment the vegetation area.

Among the pixel-based methods, one type is based on the local image gradient computed from the pixels. For example, as a feature descriptor, the Histogram of Oriented Gradient (HOG) is used to calculate the statistical value of the directional message of the local image gradient \parencite{dalal2005histograms}. In the research of automatic detection of individual oil palm trees in UAV images \parencite{wang2019automatic}, the pixels in the images were initially classified into vegetation and non-vegetation by SVM. And then, for the vegetation regions, another SVM would be used on the features extracted by HOG for detecting the oil palm tree.
Moreover, the images are scaled to different sizes for detecting oil palm trees of various sizes. Finally, the results indicated that the overall accuracy in all test sites was over 99.00 \%. Another method, the Level Sets (LSs), also used the local image gradient to sketch the outline of an object. It was applied with other methods together for tree detection in followed research.

By combining several classical algorithms, an efficient framework for palm tree detection was proposed\parencite{malek2014efficient}. There were four parts in the framework. At first, the Scale-Invariant Feature Transform (SIFT) was applied to the original images to extract the key points. SIFT can detect and match local features in images by comparing the Euclidean distance between the database features and the new image \parencite{lowe1999object}. This efficient framework indicated the feasibility of SIFT in detecting oil palm trees when it is rarely applied in this research area. 
Then, these key points were initially classified into the palm, and non-palm types by an Extreme Learning Machine (ELM), which was an excellent classification used in tree detection \parencite{kestur2018tree}. Subsequently, LSs were used to connect the points for sketching the contours of the trees. Finally, the pattern of the LSs region was further classified by the Local Binary Patterns (LBPs) to find the oil palm tree accurately. LBPs could distinguish the oil palm tree from other green plants by comparing the gray-scale invariant texture efficiently \parencite{wang1990texture}.

One of the core parts of the SIFT method is the creation of scale-space creation which is commonly applied in detection  \parencite{li2021wood}. In simple terms, the scale space is a function of and continuous with the scale parameter as the independent variable. Furthermore, it can represent the original intensity image in a multi-scale way. Another research combined the scale-space with other classical methods to detect the papaya tree with UAV images \parencite{jiang2017papaya}. At first, the RGB image was transformed to Lab color space to create a high contrast between the trees and backgrounds. The Lab color space has three channels with L for lightness, a for red-green, and b for yellow-blue. Then, the scale-space was created to generate the images with blobs symbolizing the object. To select the trees in all the objects, a local maximum filtering was applied to compare the pixel with its eight neighbors. Meanwhile, only the pixel with a value higher than a specified threshold could be classified into the tree. Finally, the results showed that the proposed method reaches a high accuracy in detecting trees as an F-score larger than 0.94.

Different from other classical methods, which are pixel-based, TM is an object-based method and has been used in tree detection \parencite{salami2019fly}. The manually built template plays a crucial role in the detection and will influence the results directly. Furthermore, the template is a standard model and will be used for comparison with the target. The target can be detected by finding the most similar region or the area with the slightest difference. Classical TM methods are Normalized with the Cross Correlation (NCC), the square root of the Sum of Squared Differences (SSD), and the Sum of Absolute Differences (SAD). In detecting oil palm trees, the researcher applied the NCC method to evaluate the agreement degree between the template tree and the actual tree in the image \parencite{hashemvand2020supporting}. 

The traditional template matching methods used in tree detection depends highly on the grey-intensity image. Such methods perform well when the image only includes trees without other interfering objects, such as buildings and cars. In order to overcome the limitations and increase the detection accuracy, one study implemented tree detection based on different vision levels including spectra, shape, and context \parencite{hung2012multi}. The detection model has two most distinctive features. The first one considers the tree shadow information during the detection. The second one, rather than only matching the grayscale intensity image, the method matches the template in the object class, and shadow class separately and then combines the results for the final decision.


\section{Conclusion and outlook}
This paper reviewed the recent application of UAVs equipped with LiDAR sensors or cameras for tree detection. The methods of tree detection can be divided into two categories based on data type, i.e., point cloud data and image data. Furthermore, the point cloud data can be collected by LiDAR sensors directly or transformed from UAV images. Since the information collected by the camera cannot penetrate the crown like the LiDAR sensor, the point cloud data contains only a minimal amount of terrain structure information. Thanks to the development of computer vision, tree detection could be conducted directly on the UAV image spectral channel. Moreover, depending on the option to use a DL-based algorithm, the methods applied to UAV images are classified as classical and DL-based. The specific conclusions derived from this review are the following:

\begin{itemize}
\item According to our statistics, the number of research based on LiDAR data kept increasing up to the year 2020 and has remained significant till now due to the high precision of LiDAR point cloud data. Meanwhile, caused by significant advances in object detection in recent years, DL-based tree detection research accounted for the most considerable portion, i.e., 45 \% of total research in 2022.

\item Whether using point cloud data or image data, the two main steps of most tree detection using classical methods are segmentation and localization or classification. Meanwhile, most DL-based tree detection methods are based on the object, meaning the detection results show the location and class of trees with only one step. Although the segmentation can show the exact area of individual trees, it is time-consuming and cannot be used in real-time detection.

\item In the detection based on point cloud data, it is mainly completed based on the following attributes: CHM, the distance between points, point density, and model represented by a minimal set of points. All methods have limitations according to the attribute, such as that CHM-based detection is unsuitable for the tree without the obvious highest point in the crown. Furthermore, the distance between points is unsuitable for tree detection when the tree branches are intertwined.

\item In the detection based on image data, the DL-based methods highly rely on the training data. Robust DL-based detection methods need to include as many tree samples as possible in the training data, e.g., under different lighting conditions, ages, health levels, and camera angles. Furthermore, the classical methods are focused on the image pixel information, such as local pixel gradient and binary image. Usually, the classical method often performs well in one specific type of tree due to the manually defined parameters.

\end{itemize}

Although extensive research on tree detection has been carried out with different types of methods, many fundamental problems still need to be solved as have been identified from our review. For the ITD in complex forest environments with multiple types of trees, real-time detection of individual trees for agricultural applications have yet to yield significant results. Therefore, the detailed analysis on the challenges and possibilities must be considered in the future study. Although the camera is an affordable alternative to the LiDAR sensor in collecting point cloud data, the LiDAR-based method is still necessary when detecting densely vegetated areas, as it can penetrate the tree canopy to obtain bottom information. Therefore, the effective ways to separate the crossed branches from the point cloud data in the densely vegetated areas is a fundamental problem of the ITD for the future.

According to our statistics, there is no doubt that the DL-based method is suitable and becoming increasingly popular in tree detection. However, the performance of DL-based detection highly relies on the architecture of the neural network and training data, which simultaneously restrict its application in different forests. Meanwhile, DL-based methods usually need more pertinence, i.e., detecting the tree with unique properties does not achieve good results. Therefore, how to make DL methods more general to be applied to a region containing multiple trees, such as tropical rainforest, is a question that people still need to consider in the future. 

The classical method has yet to attract much attention due to its high dependence on manually set parameters in recent years. Because of such characteristics, the classical methods could be a promising solution for tree detection with specific features. Meanwhile, due to the possibility of building a classical method with a simple structure and high efficiency, real-time tree detection could be achieved in agricultural applications, such as precision spraying, pollination, and harvest.

\section*{Acknowledgement} 
The authors would like to thank Malaysian Ministry of Higher Education (MOHE) for providing the Fundamental Research Grant Scheme (FRGS) (Grant number: FRGS/1/2020/TK0/USM/03/3).

\printbibliography
\end{doublespace}

\end{document}